\documentclass{article} 
\usepackage[final]{colm2025_conference}

\usepackage{microtype}
\usepackage{hyperref}
\usepackage{url}
\usepackage{booktabs}

\usepackage{lineno}
\usepackage{multirow}
\usepackage{graphicx}
\usepackage{amsmath}
\usepackage{enumitem}
\usepackage{textcomp}
\usepackage{xspace}
\usepackage{csquotes}
\definecolor{darkblue}{rgb}{0, 0, 0.5}
\hypersetup{colorlinks=true, citecolor=darkblue, linkcolor=darkblue, urlcolor=darkblue}

\usepackage{graphicx}
\usepackage{multirow}
\usepackage{pifont}
\usepackage{colortbl}
\definecolor{best_green}{HTML}{78B840}
\definecolor{light_green}{HTML}{ADF76D}
\definecolor{mid_yellow}{HTML}{F7F06D}
\definecolor{bad_orange}{HTML}{FABD43}
\definecolor{bad_red}{HTML}{F2300A}
\usepackage{amssymb}
\usepackage{hyperref}

\title{Decoding Alignment: A Critical Survey of LLM Development\\ Initiatives through Value-setting and Data-centric Lens}

\author{Ilias Chalkidis\thanks{As this is a single-author paper, the use of the word `we' shall be interpreted as inviting you, the reader, to follow the thought process, the survey, and the documentation of findings and concerns. }\\
Department of Computer Science, University of Copenhagen, Denmark\\
\texttt{[firstname].[lastname][at]di.ku.dk}
}

\begin{document}

\ifcolmsubmission
\linenumbers
\fi

\maketitle

\begin{abstract}
AI Alignment, primarily in the form of Reinforcement Learning from Human Feedback (RLHF), has been a cornerstone of the post-training phase in developing Large Language Models (LLMs). It has also been a popular research topic across various disciplines beyond Computer Science, including Philosophy and Law, among others, highlighting the socio-technical challenges involved. Nonetheless, except for the computational techniques related to alignment, there has been limited focus on the broader picture: the scope of these processes, which primarily rely on the selected objectives (values), and the data collected and used to imprint such objectives into the models. This work aims to reveal how alignment is understood and applied in practice from a value-setting and data-centric perspective. For this purpose, we investigate and survey (`audit') publicly available documentation released by 6 LLM development initiatives by 5 leading organizations shaping this technology, focusing on proprietary (OpenAI's GPT, Anthropic's Claude, Google's Gemini) and open-weight (Meta's Llama, Google's Gemma, and Alibaba's Qwen) initiatives, all published in the last 3 years.\footnote{The aim of naming the tech corporations involved is to make it clear from the get-go that this is predominantly, and in the case of this work, exclusively, a corporate environment.} The findings are documented in detail per initiative, while there is also an overall summary concerning different aspects, mainly from a value-setting and data-centric perspective. On the basis of our findings, we discuss a series of broader related concerns.\\
\noindent\makebox[\linewidth]{\rule{10cm}{1pt}} 
\underline{\textbf{Note}}: This is a working paper and will be updated with new information or corrections based on community feedback. Update: \today \\
\end{abstract}
\vspace{-7mm}

\section{Introduction}

Large Language Models (LLMs) exhibit impressive capabilities~\citep{achiam2023gpt,llama3,team2025gemini}, excelling in a plethora of instruction-following tasks such as open-domain question answering, solving arithmetic and coding tasks, providing suggestions, creative writing, and more~\citep{bommasani2023holistic, chiang2024chatbot}. Ensuring that these highly capable AI systems behave in ways consistent with user intentions and developer-designed policies, \emph{AI Alignment} has consequently become paramount in LLM development initiatives. Within the technical pipeline of LLM development, Reinforcement Learning from Human Feedback (RLHF), or other similar techniques, has emerged as the de facto cornerstone methodology for the post-training alignment phase. By fine-tuning models based on human preferences, RLHF aims to steer the behavior of the models, as manifested via model responses, towards the desired objectives\footnote{In this work, the terms `value' and  `objective' are used interchangeably. The term `value' is often used as a weasel word in the literature to misportray mostly narrow service-oriented objectives as universal `human values'~\citep{askell2021general, hendrycks2021aligning, claude3}.} inferred by the human-labeled datasets.

The significance of AI alignment extends far beyond the confines of Computer Science. Its implications resonate deeply across a spectrum of disciplines, including Law~\citep{caputo2024alignment,caputo2024rulescasesreasoningpositivist}, Philosophy~\citep{gabriel2020artificial, christian2020alignment, gabriel2025matter}, Ethics~\citep{dignum2019responsible, muller2020ethics} in particular, and the Social Sciences~\citep{artz2023machine,kirk2025human,jordan2025collectivist}, among others, where researchers grapple with questions of value specification, societal impact, and accountability. This cross-disciplinary interest underscores AI alignment's role not merely as a technical challenge meant to be solved by computer scientists, but as a socio-technical challenge that involves both technical, as well as normative decisions~\citep{gabriel2020artificial} involving political and ethical considerations.

\subsection{What are LLMs and AI alignment (in a nutshell)?}

\emph{(This section is mainly aimed for readers with no technical background, or readers that have not followed closely recent advances related to LLMs post-2018.)}

LLMs are large (billion-parameters-sized) Transformer-based~\citep{Vaswani2017} language models, initially pre-trained as standard \emph{language models}~\citep{bengio-lm-2000,radford2018improving}, i.e., next-word-predictors, in vast web textual corpora. By learning the language's inner workings (distribution), pre-trained language models can generate fluent (natural) text and compress and memorize high volumes of knowledge, which on its own is proven to be a solid backbone to rely on building Natural Language Processing (NLP) models for specific downstream tasks~\citep{radford2018improving, devlin-etal-2019-bert}. Seeking the fascinating idea of general-purpose models, which can solve and generalize over many different tasks, PLMs are further optimized to follow user \emph{instructions}~\citep{chung2022scaling} in the form of question-answer pairs for diverse downstream tasks, e.g., question-answering, solving arithmetic and coding tasks, providing suggestions, summarizing long documents, creative writing, etc. Nonetheless, instruction-tuned LLMs still lack in terms of generalization, understanding user intentions, and conforming to user expectations.

To do so, the training pipeline is further augmented, including an additional step of the so-called \emph{AI alignment}~\citep{leike-etal-2018-alignnment, ouyang2022training, bai2022training} with Reinforcement Learning from Human Feedback (RLHF) \citep{christiano-etal-2017-drl,Stiennon2020} or similar techniques.\footnote{In the scope of this survey, \emph{AI alignment} refers to the alignment of LLMs. AI alignment at large follows similar principles and techniques. What makes alignment a particular challenge for LLMs is that objective functions cannot be formalized to shape well-defined rewards, a general challenge for language generation, so human feedback is necessary to assess the quality of responses of LLMs.} In this step, the model is aligned to (optimized for) a set of predetermined objectives, e.g., being helpful, avoiding the use of harmful language, denying illegal requests, not spreading misinformation, etc. This process heavily relies on utilizing human ratings (labels) over alternative model-generated responses that early versions of instruction-tuned LLMs produce, given a user instruction. This is a contrastive setup between preferred vs. non-preferred (rejected) responses.

More recently, AI alignment, originally relying on large-scale human-curated data, is increasingly relying on \emph{synthetic (model-generated) data}, referred to as Constitutional AI~\citep{bai2022constitutional}, or more broadly defined as Reinforcement Learning from AI Feedback (RLAIF). The main idea is that, considering the high steerability of LLMs, one can prompt a model to respond based on a set of predetermined stated objectives, enforcing the generation of curated responses that will later be used in contrast to unfiltered (non-curated) responses to further fine-tune the model to adopt the desired behavior in a similar setup with RLHF. 

With the rise of \emph{reasoning} LLMs~\citep{openai2024openaio1card,guo2025deepseek}, that learn to generate chain-of-thoughts~\citep{wei2022cot} as a preface to solving complex tasks, the use of deliberative alignment~\citep{guan2025deliberative} has emerged as a prominent technique to align the models' reasoning without requiring human-written chain-of-thoughts; in the same fashion as RLAIF, i.e., the use of curated synthetic data (responses including chain-of-thoughts) in contrast to non-curated ones.

\subsection{Scope}

Despite this widespread recognition and the intense focus on AI alignment, a critical knowledge gap persists. While the computational and optimization aspects of RLHF, i.e., how optimization algorithms~\citep{christiano-etal-2017-drl,rafailov2023direct,shao2024deepseekmath} utilize human feedback from preference data, have received considerable attention, our understanding of the broader process remains remarkably opaque and sterilized. 

This work argues that the actual scope and substance of alignment achieved via RLHF or other techniques are determined less by the intricacies of the optimization algorithms themselves, and far more by foundational choices concerning:

\textbf{Value Specification}: What objectives are selected as targets for alignment? Who defines these objectives, and through what process? How are they operationalized? How are these objectives prioritized when in conflict?

\textbf{Data Curation and Application}: What data (human preferences or other labeling) is collected to represent these objectives?  How are data annotators involved in this process? How is this data sourced, filtered, and utilized to "imprint" the chosen values onto the model?

These choices are inherently normative and contextual, yet they are often shrouded in vagueness and industrial secrecy or described at a high level, if not at all. Hence, the aim is to document how alignment-focused processes are concretely understood, scoped, and implemented in practice by the leading organizations shaping this technology, to the best of our abilities, given the available documentation (articles, tech reports, system cards, etc.). Importantly, we want to track how the design choices and processes developed over time. Once we chart the landscape, we can consider limitations and raise critical concerns.

\subsection{Contributions}

This work explores AI alignment in the context of Large Language Model (LLM) development. To do this, we first establish a framework to analyze alignment-focused processes from a value-setting and data-centric perspective in Section~\ref{sec:methodology}. Using this framework, in Section~\ref{sec:documentation_analysis}, we survey documentation from 6 LLM initiatives over the past 3 years, regarding proprietary and open-weight models, with the analysis revealing important findings.

Our survey shows that developers universally define AI alignment around three core pillars: helpfulness, harmlessness, and truthfulness. However, the decision-making authority behind these objectives is often unclear, with developers--acting as a proxy for tech companies--making these choices with limited or no reflection, while we also find that developers rarely specify how to prioritize these objectives when they conflict. When it comes to data collection processes, information about the annotation framework for preference data is mostly transparent. In contrast, details on how annotators are selected, trained, and how their backgrounds might influence the data they collect are limited. While the use of data in fine-tuning models is mostly well-documented, details on prompt selection and data filtering are not.

The aforementioned findings raise several concerns, discussed in Section~\ref{sec:concerns}. A primary issue is the lack of transparency in recent work, particularly regarding who determines alignment objectives and the need for broader considerations to escape corporate alignment. The field has yet to move beyond the primitive Helpfulness-Harmlessness-Honesty (HHH) framework outlined by~\cite{askell2021general}, suggesting a stagnation in recent approaches. We also discuss how the non-disclosure of human feedback data collection in recent work raises serious questions about labor obfuscation. Concurrently, we note that over-reliance on AI, as evidenced by recent studies, could lead to cognitive offloading and a subsequent deskilling of users. Furthermore, we discuss how the examined work neglects broader political, geopolitical, and cultural biases, and emphasize how models demonstrate clear Western hegemonic ideological biases. Lastly, we consider the consequential "flattening effect" that alignment may induce under the current paradigm, which could lead to a significant further diminishment of cultural pluralism.

\section{Methodology}
\label{sec:methodology}

\subsection{Defining Points of Interest}
\label{sec:points_of_interest}

To address the question: ``How AI alignment is understood and applied in practice for LLM development?'', we have to establish what exactly we are examining. Hence, we identify \emph{points of interest} through an exercise, envisioning how alignment-focused processes are orchestrated. The development sketch is based on published accounts of early initiatives~\citep{askell2021general,ouyang2022training, bai2022training} and is informed by established norms and expectations around high-quality data collection in machine learning workflows.

\subsubsection{Sketch of the data-centric alignment-focused process}
\label{sec:sketch}

We outline a high-level sketch of the alignment-focused process, dividing it into three phases: pre-hoc, ad-hoc, and post-hoc data collection, considering the value-setting and data-centric focus of our study. Each phase reflects a cluster of related activities that shape how alignment is operationalized, from early conceptual design through to model development, deployment, release, and potential data sharing.

\paragraph{A. Pre-Hoc -- Planning Phase}

(i) \emph{Specification of Objectives}: First, the development team defines the objectives of the alignment process, i.e., the ``behaviors'' models are expected to learn. Key considerations include who determines these objectives, why these particular objectives are chosen, i.e., what the selection criteria are, and how precisely they are defined. Objectives may range from narrowly specified to broadly defined. In more open-ended cases, alignment may rely on the idea of merely aggregating diverse human preferences under very limited or no specifications, leveraging the ``wisdom of the crowd''.

(ii) \emph{Specification of the Annotation Approach}: Next, the team determines the most appropriate annotation strategy, specifying how human feedback will be collected, e.g., through comparisons, edits, or labels on model-generated responses, what the annotation interface looks like, and how the overall approach aligns with project constraints such as time and budget. The selection of approach depends on the anticipated quality and nature of the data signal to be captured, which is connected to the downstream usage of the collected data in follow-up experiments.

(iii) \emph{Annotator Selection and Preparation}: This step involves selecting and preparing annotators based on some predefined criteria. The development team may conduct training sessions, assess annotator performance through pilot phases, and implement screening procedures to identify suitable candidates. Relevant factors include the recruitment strategy, i.e., internally or externally via vendors, consideration or not of demographics, development of training materials, e.g., guidelines, and evaluation protocols. The level of rigor can vary significantly from thorough onboarding to minimal instructions, depending on the design philosophy. Similarly, considerations of the annotators' demographics may or may not be a priority.

\paragraph{B. Ad-Hoc -- Data Collection Phase}

(i) \emph{Data Annotation and Process Review}: During this step, data is actively collected and annotated. The development team oversees the annotation workflow, if possible, and potentially monitors data quality, and iteratively refines the process. Feedback loops and adjustments may lead to revisions of prior decisions (pre-hoc phase), reflecting an incremental development model. Again, the depth of reflection and refinement is connected with project constraints such as time, budget, and control.

\paragraph{C. Post-Hoc -- Model Fine-tuning Phase}

(i) \emph{Data Filtering}: (Optional) Collected data may be filtered post-hoc to enforce quality assurance standards, e.g., excluding outliers based on diagnostic criteria (lengthy responses, violation of policies, etc.), or manual reviewing.

(ii) \emph{Use of Data}: The development team uses the curated data to optimize models in a post-training phase. This may involve training reward models for RLHF or directly fine-tuning the base model to better align with the predefined objectives as reflected by the data.

(iii) \emph{Publication of Data} (Optional) The team may choose to publish the dataset, or a subset thereof, to encourage knowledge sharing and facilitate further research following open science principles. This step is optional and may be omitted, considering that data is the intellectual property of the corporation that provides a competitive edge.

\subsubsection{Remarks}

The aforementioned exercise can be incomplete or partially differ from the reality of the development projects that we will examine. Nonetheless, it is important to set considerable expectations of \emph{what \underline{we would like to know}} following a critical perspective and not simply ``fall'' for \emph{what \underline{they (developers/corporations) would like to share}} irrespective of the reasons, e.g., underestimating information sharing and open science, industrial secrecy to keep a competitive edge, neglect of accountability, etc.. 

\begin{table*}[t]
    \centering
    \begin{tabular}{cc|c|c|c|c}
         \multicolumn{2}{c|}{\textbf{Initiative / Project}} & \multicolumn{4}{c}{\textbf{Details}} \\
          \textbf{Company} & \textbf{Name} & \textbf{Status} & \textbf{Version} & \textbf{Release Date} & \textbf{Documentation}  \\
         \midrule
         \multirow{4}{*}{OpenAI} & \multirow{4}{*}{GPT} & \multirow{4}{*}{Closed} & 3.5 & Mar 2022 & \citet{ouyang2022training}\\
          & & & 4 & Nov 2023 & \cite{achiam2023gpt} \\
          & & & 4.5 & May 2024 & \citet{hurst2024gpt}\\
          & & & 5 & Aug 2025 & \citet{openai2025gpt5} \\
          \midrule
          \multirow{4}{*}{Anthropic} & \multirow{4}{*}{Claude} & \multirow{5}{*}{Closed} & 1* & Mar 2023 & \citet{bai2022training, bai2022constitutional} \\
          & & & 2 & Jul 2023 & \citet{claude2}\\
          & & & 3/3.5 & Mar/Jun 2024 & \citet{claude3} \\
          & & & 3.7 & Feb 2025 & \citet{claude37} \\
          & & & 4 & May 2025 & \citet{claude4} \\
          \midrule
          \multirow{4}{*}{Google} & \multirow{4}{*}{Gemini} & \multirow{3}{*}{Closed} & 1 & Feb 2024 & \citet{team2023gemini} \\
          & & & 1.5 & Mar 2024 & \citet{team2024gemini15} \\
          & & & 2 & Feb 2025 & n/a \\
          & & & 2.5 & Mar 2025 & \citet{team2025gemini} \\
          \midrule
          \multirow{4}{*}{Meta} & \multirow{4}{*}{Llama} & \multirow{4}{*}{Open} & 1 & Feb 2023 & \citet{touvron2023llama1}\\
          & & & 2 & Jul 2023 & \citet{touvron2023llama2} \\
          & & & 3/3.1 & Jul 2024 & \citet{llama3}\\
          & & & 4 & Apr 2025 & \citet{llama4} \\
          \midrule
          \multirow{4}{*}{Google} & \multirow{4}{*}{Gemma} & \multirow{3}{*}{Open} & 1 & Jun 2024 & \citet{team2024gemma1}\\
          & & & 2 & Jul 2024 & \citet{team2024gemma2} \\
          & & & 3 & Mar 2025 & \citet{team2025gemma3} \\
          \midrule
          \multirow{4}{*}{Alibaba} & \multirow{4}{*}{Qwen} & \multirow{3}{*}{Open} & 1 & Sep 2023 & \citet{qwen1technicalreport} \\
          & & & 2 & Sep 2024 & \citet{qwen2technicalreport} \\
          & & & 2.5 & Jan 2024 &\citet{qwen25technicalreport} \\
          & & & 3 & May 2025 & \citet{qwen3technicalreport} \\

    \end{tabular}
    \vspace{-2mm}
    \caption{LLM development projects split into initiatives alongside the status (closed, or open-weight), the release date, and the most relevant documentation. *We account the work of ~\cite{bai2022training,bai2022constitutional} as the main work related to the follow-up release of Claude 1, since there is no other documentation available.}
    \vspace{-2mm}
    \label{tab:models}
\end{table*}

\subsection{``Auditing'' via Documentation}
\label{sec:questionaire}

Following the sketch, we design a questionnaire regarding the the points of interest, and proceed in surveying publicly available documentation, comprising by academic-style articles, tech reports, and model/system cards, that has been released by 6 major LLM development initiatives, focusing on both proprietary (OpenAI's GPT, Anthropic's Claude, Google's Gemini) and open-weight (Meta's Llama, Google's Gemma, and Alibaba's Qwen) initiatives.\footnote{In this article, the work from Mistral AI~\citep{jiang2023mistral,jiang2024mixtral,mistralai2025magistral} is not examined, since the available documentation do not discuss alignment-focused practices in any meaningful way, especially in the context of our value-setting and data-centric focus, apart from a broad statement in~\cite{jiang2024mixtral}: ``\emph{We train Mixtral–Instruct using [...] Direct Preference Optimization (DPO) on a paired feedback dataset.}'', while \cite{mistralai2025magistral} states that for Magistral, the scope of alignment is limited to problems with verifiable solutions. Similarly, other recent projects such as DeepSeek~\citep{guo2025deepseek}, OLMO~\citep{Groeneveld2023OLMo}, Kimi~\citep{kimiteam2025kimik2openagentic} are not considered.} In Table~\ref{tab:models}, we see the examined initiatives (model series) alongside the related documentation. This survey should be understood as a form of--by no means ideal--AI auditing~\citep{birhane2024aiauditingbrokenbus}, as the goal is to understand and reveal information on the practices related to alignment-focused processes conducted -predominantly, and in the case of this study exclusively- by private tech corporations. It involves looking for ``a needle in the haystack'', since information is sometimes ``buried'' in the appendices of 100-page-long documents, sitting behind private domains, or cross-referencing other documents.

\subsubsection{Questionnaire}

Based on the developed sketch (Section~\ref{sec:points_of_interest}), we compile a questionnaire of 10 broad questions that we aim to answer for each examined initiative:

(1) \textbf{What are the selected objectives that developed models should align with?} We want to identify clearly stated objectives. We look for statements that show intent rather than vague references to how the model may or may not improve, considering a specific aspect.

(2) \textbf{Who decided on the appropriateness of the selected objectives?}  We want to identify who is the ``respective authority'' deciding the selected objectives, e.g., the development team, the tech corporation (employer), or an external independent authority/organization.

(3) \textbf{How are the selected objectives prioritized when they conflict?} We want to identify how the selected objectives may conflict and be prioritized in the form of instructions (guidelines) to the annotators, primarily, or post-hoc mitigation techniques, e.g., weighting data mixtures, multi-loss functions, etc.

(4) \textbf{How is the data annotation process designed?} We want to identify how the annotation process is designed, i.e., what form of data the developers intend to collect related to human feedback, e.g., comparison between model-generated responses, edits, and other forms of labeling (tagging) as metadata.

(5) \textbf{What is the process of hiring, selecting, and preparing annotators?} We want to identify information related to the hiring process, e.g., internal or external via vendors, and the selection of annotators considering pilots and screening, alongside the intention for matching specific demographics, and the number of individuals involved. Importantly, we want to identify information related to the training of annotators given specific guidelines.

(6) \textbf{What is the process for selecting the prompts (used to generate model responses and gather user preferences)?} We want to identify how the pool of prompts used for data annotation processes, i.e., rating/labeling, is compiled. This is related to how the developers intend to prioritize and diversify the data collection across topics or other considerations.

(7) \textbf{Are the collected human preference data filtered?} We want to identify, in case any filtering is applied, i.e., exclusion of data samples, what the related criteria and practices are to enforce potential quality assurance standards in a post-hoc manner. 

(8) \textbf{What is the volume of the collected human preference data?} We want to identify the volume of the data collected related to alignment-focused processes. We want to get an idea of the scale of the data needed, i.e., are these thousands or millions of data samples?

(9) \textbf{How are the collected human preference data used to fine-tune (align) the models?} We want to identify how the collected data are used to instill the desired ``behaviors'' (objectives)--as captured by the data--in the developed models.

(10) \textbf{Are the collected human preference data being published?} We want to identify if the collected data, or parts, are published. We're interested on how companies view this data, as a public resource that can be openly shared and examined, or as their private property.

\section{Documentation of Findings}
\label{sec:documentation_analysis}

In Section~\ref{sec:per_initiative}, we first review the results grouped by initiative (model family), e.g., collectively for all the Meta's Llama models, rather than individually. The main reason is that, as new models are released and their capabilities get enhanced, the related documentation concentrates less on formally introducing and describing the alignment-focused processes, and mainly presents more benchmarking results, red-teaming efforts, and new capabilities, e.g., processing and generating images, and other modalities, or using tools, etc. Nonetheless, our goal is also to track how the design choices and processes evolved over time in the lifecycle of a given model series. In Section~\ref{sec:overall_findings}, we consider the overall findings.

\subsection{Findings per Initiative}
\label{sec:per_initiative}

\emph{(The specific section is long and the information is quite repetitive across initiatives. Readers who want to get a grasp of the bigger picture can skip to Section~\ref{sec:overall_findings}, where we review the overall findings.)}

The findings are structured in a Q\&A format, following the developed questionnaire (Sections~\ref{sec:questionaire}). We quote parts from the documentation (in the form ``\emph{[...]}'') extensively to preserve verbatim content and present the findings faithfully to the source, as much as possible. If information related to a specific project (release) is not mentioned, it can be implied that such information was not available (N/A).

\subsubsection{OpenAI's GPT} 

\textbf{What are the selected objectives that developed models should align with?} 

In \cite{ouyang2022training} describing the development of InstructGPT 3.5, the authors clearly state that they use ``\emph{a framework similar to \citet{askell2021general}, who define models to be aligned if they are helpful, honest, and harmless.}''. As we see later on, \emph{helpfulness}, \emph{harmlessness}, and \emph{truthfulness} (also coined as honesty), are the three pillars of AI Alignment (Figure~\ref{fig:objectives}) in almost every LLM development project.\footnote{We will refer to \citeauthor{askell2021general}'s foundational `Helpfulness-Harmlessness-Honesty' (HHH) framework, as HHH in short, in the rest of the paper.} 
\citeauthor{ouyang2022training} state regarding helpfulness: ``\emph{To be helpful, the model should follow instructions, but also infer intention}''. When it comes to honesty, the authors are sceptical about how it can be operationalized: ``\emph{It is unclear how to measure honesty [...]; since the model is a big black box, we can’t infer its beliefs.}'', and focus instead on measuring truthfulness: ``\emph{whether the model’s statements about the world are true}''. The authors also discuss the difficulty of operationalizing harmlessness: ``\emph{[...] In most
cases, the harms from language models depend on how their outputs are used in the real world. [...] Therefore, we use a suite of more specific proxy criteria that aim to capture different aspects of behavior [...] that could end up being harmful: [...] labelers evaluate whether an output is inappropriate in the context of a customer assistant, denigrates a protected class, or contains
sexual or violent content.}''. For GPT-4~\citep{achiam2023gpt}, the authors mention their intent for  \emph{"increased helpfulness and harmlessness"}, while they also state their intent towards increased factuality and less hallucinations, related to truthfulness. When it comes to harmlessness and safety, the authors mention their aim to mitigate disinformation, privacy risks, cybersecurity, and the proliferation of weapons. For GPT4.5~\citep{hurst2024gpt}, the authors vaguely mention: ``\emph{Safety training for political persuasion tasks.}''. For GPT5~\citep{openai2025gpt5}, the authors mention: ``\emph{we post-trained our models to reduce sycophancy.}'', and further vaguely mention: ``\emph{We trained [our models] to follow OpenAI’s safety policies}''. There is also mention of reducing hallucinations.

\textbf{Who decided on the appropriateness of the selected objectives?} 

In \cite{ouyang2022training}, the authors make clear that they decided on the objectives: ``\emph{our preferences, as the researchers designing this study (and thus by proxy to [...], OpenAI)}'', while earlier mention that the model aligns with ``\emph{stated preferences of a specific group of people [...], rather than any broader notion of “human values”}''. The authors briefly discuss related challenges of designing an alignment process, considering a broader consensus in Section 5.4 "Open questions".  In a similar tone, for GPT-4, \citet{achiam2023gpt} mentions: ``\emph{It is also likely that our approach to sourcing researchers privileges the kinds of risks that are top of mind in academic communities and at AI firms.}''. The developers are well aware that the objectives are defined solely based on what they, as members of the organization, consider important, which does not account for broader legitimacy. Late work~\citep{openai2025gpt5} refers to OpenAI as the immediate authority.

\textbf{How are the objectives prioritized when they conflict?}  

In \cite{ouyang2022training}, the authors mention: ``\emph{[..] during the labeling of our training data, we had labelers prioritize helpfulness to the user as the most important criteria [...], whereas in our final evaluations we had labelers prioritize truthfulness and harmlessness.}''. For GPT-4~\citep{achiam2023gpt} and follow-up work, there is no mention of objective prioritization.

\textbf{How is the data annotation process orchestrated?} 

In \cite{ouyang2022training}, the authors follow a two-step approach, where they ``\emph{present labelers with anywhere between K = 4 and K = 9 responses [...]}''. First: ``\emph{For each output [alternative model response], labelers give a Likert score for overall quality on a 1-7 scale, and also provide various metadata labels.}''; inspecting the screenshot of the labeling interface the labelers have to answer for each model response individually, if the model response follows the user instruction, is inappropriate for customer assistance, gives harmful advice, express moral judgment,  denigrates a protected class, contains sexual/violent content. Then, labelers had to rank (from best to worst), while ``\emph{Ties are encouraged in cases
where two outputs seem to be of similar quality.}''. For GPT-4~\citep{achiam2023gpt}, the authors refer to their prior work and describe the same process briefly.

\textbf{What is the process of hiring, selecting, and preparing annotators?} 

In \cite{ouyang2022training}, the authors mention: ``\emph{Our labelers consist of contractors hired either through Upwork, or sourced from Scale AI}'', and mention 40 contractors (labelers) in total. Concerning screening: ``\emph{we conducted a screening process to select labelers who showed a high propensity to detect and respond
to sensitive content.}''. The screening primarily considers the agreement between the labelers and developers regarding sensitive content flagging and preference rankings, while also selecting a diverse group that can accurately identify sensitive speech. The developers curate guidelines that evolved throughout the project based on feedback and improved understanding of the process\footnote{The authors provide a link to the final version of the guidelines, available \href{https://docs.google.com/document/d/1MJCqDNjzD04UbcnVZ-LmeXJ04-TKEICDAepXyMCBUb8/edit?usp=sharing}{here}.} The authors provide demographics; the labelers are English-speaking, based predominately in SE Asia (53\%) and U.S (17\%). The labelers identify almost 50/50 male and female, with 5\% non-binary or other. The vast majority (75\%) is under 35 years old, with 90\% having an undergraduate or master's education. For GPT-4~\citep{achiam2023gpt} and follow-up work, there is no further information.

\textbf{What is the process for selecting the prompts?} 

In~\cite{ouyang2022training}, the authors mention: ``\emph{Our prompt dataset consists primarily of text prompts submitted to the OpenAI API [...] on the Playground interface}''. The authors mention that they heuristically deduplicate prompts, limit the number of prompts per user, and filter out prompts that include personally identifiable information (PII). For GPT-4~\citep{achiam2023gpt} and follow-up work, there is no further information.

\textbf{Are the collected human preference data filtered?} 

In~\cite{ouyang2022training}, the authors mention: ``\emph{We train all the RL models [...] after filtering out prompts with PII and deduplication based on common prefixes.}'', while also mention later: ``\emph{We filter out prompts that are longer than 1k tokens and limit the maximum response length to 1k tokens.}''. For GPT-4~\citep{achiam2023gpt}, there is no further information.

\textbf{What is the volume of the collected human preference data?} 

In~\cite{ouyang2022training}, the authors mention that they collected human preference data for approx. 10K prompts from the hired labelers, and approx. 41K prompts from OpenAI customers (Table 6). It is unclear how preference (comparison) data were collected from OpenAI customers without proper guidelines and onboarding. For GTP-4, \cite{achiam2023gpt} mentions: ``\emph{we [...] collect additional labeled comparison data that we use to train our reward models.}'', without further information. The authors also mention the use of synthetic comparison data on closed-domain hallucinations, which: ``\emph{we also mix into our RM dataset}''. For GPT-4~\citep{achiam2023gpt} and follow-up work, there is no further information.

\textbf{How are the collected human preference data used to fine-tune (align) the models?} 

In~\cite{ouyang2022training}, the authors mention: ``\emph{we train [our Reward Model (RM)] on all $\binom{K}{2}$ comparisons from each prompt as a single batch element}'', in other words the $K$ ranked responses are grouped into pairs of comparisons, i.e, preference of one response over another, where ties are dropped. The authors do not mention any special use of the fine-grained ranking (1-5), nor the additional metadata (labeling) related mostly to safety concerns. For GPT-4~\citep{achiam2023gpt} and follow-up work, there is no further information.

\textbf{Are the collected human preference data being published?} 

There are no available published data related to OpenAI's GPT series.\footnote{The only publicly available data comes from earlier projects loosely related to OpenAI's GPT initiative; specifically, a dataset of 179K binary comparisons related to summarization from \cite{Stiennon2020}, and 16.9K comparisons for long-form question-answering from \cite{nakano2021webgpt}.}

\subsubsection{Anthropic's Claude}

\textbf{What are the selected objectives that developed models should align with?} 

In the work of \cite{bai2022training}, the closest work to the Claude 1 model,  the authors mention: ``\emph{We would like [...] to train AI agents that are helpful, honest, and harmless}'' following~\cite{askell2021general}. Nonetheless, they do not strictly define these objectives and, as they mention: ``\emph{for the most part we simply let our crowdworkers interpret these concepts as they see fit.}''. As part of this decision, they refer to their hope for data diversity and the ``wisdom of the crowd''. For Claude 2~\citep{claude2}, the authors restate that the core research focus is concentrated training models that are helpful, honest, and harmless. In this direction, the authors refer to the major impact of the work of~\cite{bai2022constitutional} to steer models and generate synthetic data using AI feedback based on Claude's constitution,\footnote{\url{https://www.anthropic.com/news/claudes-constitution}} presenting as they mention: ``\emph{explicit values [...], rather than values determined implicitly via [...] human feedback.}''.\footnote{Claude's constitution includes statements that rely on the Universal Declaration of Human Rights (UDHR), which aims to promote freedom, equality, liberty, non-discrimination, fairness, respectfulness, participation, and the right to private life, while it opposes torture, cruelty, and degrading treatment. The constitution also includes statements that encourage the consideration of non-Western perspectives, ethical and moral awareness, among others, which can be mainly reduced to the HHH framework.} The authors also state: ``\emph{Claude 2 has been trained to generate coherent documents of up to [...] roughly 3000 words''}. For Claude 3~\citep{claude3}, the authors repeat their intention to follow the HHH framework and Claude's constitution as the driving force of alignment, while there is a special notice on Claude's constitution addition on encouraging respect for disability rights.\footnote{This addition most likely comes from the work of \cite{huang2024collective} -collaboration with Anthropic-, where the human study group expressed that AI systems should be adaptable/flexible for, and understanding of, people with disabilities.} Follow-up work~\citep{claude37,claude4} (Claude 3.7/4) offers no new information.

\textbf{Who decided on the appropriateness of the selected objectives?} 

In the work of \cite{bai2022training}, the HHH framework comes from~\cite{askell2021general}--affiliated with Anthropic--, and is applied by the developers, who argue that: ``\emph{This question should not be the provenance of researchers only.}'', and comment on the potential of ``\emph{an independent organization with ethical, legal, and cultural expertise}'' to oversee such processes and curate related datasets. Nonetheless, the authors present their work in the context of ``\emph{Alignment with Human Values}''. For Claude 3~\citep{claude3}, the authors state: ``\emph{We take [...] steps [...] drawing on guidance from the NIST AI Risk Management Framework.}''.\footnote{\url{https://www.nist.gov/itl/ai-risk-management-framework}} In the work of \cite{claude3, claude37} reporting Claude 3 and 3.7, the authors state: ``\emph{Anthropic used [...] Constitutional AI [...] to align Claude with human values}''. Anthropic's work seems to be the most persistent across all initiatives in the framing of ``alignment with human values''.

\textbf{How are the objectives prioritized when they conflict?} 

\cite{bai2022training} does not discuss any guidelines related to objective prioritization.\footnote{Potentially, because they curate two separate datasets for the two main objectives: helpfulness, and harmlessness. Although this does not imply by any means that objectives cannot be at odds.}  Although they acknowledge the tension between helpfulness and harmlessness, which they try to mitigate by re-weighting the impact of the two objectives in a post-hoc manner, since they collected data for the two objectives separately (see below).

\textbf{How is the data annotation process orchestrated?} 

\cite{bai2022training} mentions: ``\emph{[Labelers] can interact
with our models in natural language via chat,[...] When it's the model’s conversational turn, users see two possible model responses and choose one with which to proceed.}''. In this case, we have binary comparisons instead of ranking among many alternative responses, as ~\cite{ouyang2022training} did, while we also have multi-turn chat discussions. The authors conducted annotation projects for helpfulness and harmlessness separately. The authors included: ``\emph{an additional option [..] feature allowed them to edit one of the model responses}''. They also allow labelers to express a preference strength. For harmlessness, the authors curate a red-teaming effort where: ``\emph{[they] asked crowdworkers to attempt to elicit harmful responses from [their] models, and to choose the more harmful response}''. In other words, the labelers had to ``trigger'' the model to account for such scenarios. The authors provide screenshots for the guidelines, where objectives are explained, alongside some positive and negative examples.

\textbf{What is the process of hiring, selecting, and preparing annotators?} 

In~\cite{bai2022training},  the authors mainly hired master-qualified US-based MTurk workers, screening ``\emph{those who were most prolific}'' and also mention: ``\emph{We did not filter workers based on agreement or other direct measures of label quality}''. They also later hired crowdworkers on Upwork following a similar onboarding process. The labelers are solely US-based, predominantly White (68\%), and are almost equally split among male/female, with 10\% identifying as LGBTQ+. The vast majority (approx.~80\%) is 45 years old or younger, with 70\%+ having undergraduate or master's education. For Claude 3, ~\cite{claude3} states: ``\emph{Anthropic works with several data work platforms which are responsible for engaging and managing data workers who work on Anthropic’s projects. Data work tasks include selecting preferred model outputs [...] to train AI models to align with those preferences; evaluating model outputs according to a broad range of criteria [...] and adversarially testing [...] to identify potential safety vulnerabilities.}''. For Claude 4~\citep{claude4}, the authors include a similar statement.

\textbf{What is the process for selecting the prompts?} 

In~\cite{bai2022training},  the authors mention: ``\emph{Crowdworkers write a chat message to our models, asking them to perform a task, answer a question,
or discuss any topic of interest.}'', in which case we can infer that the prompts are collected on-the-fly, and there is no prior pool of prompts to select from. Later work has no further information.

\textbf{What is the volume of the collected human preference data?} 

In~\cite{bai2022training},  the authors do not explicitly state the volume of the collected data, although they published their preference datasets (see below), which account for approx.~160K pair comparisons. They also mention: ``\emph{Our work has benefited from other publicly available alignment-related data}'', such as the data from the work of ~\cite{Stiennon2020} (approx.~ 180K pairs), but there are potentially more. For Claude 2~\cite{claude2}, the authors mention: ``\emph{We have also integrated some non-English human feedback data into our process.}'', but there is no information on the scale of the collected data for any follow-up work.

\newpage

\textbf{Are the collected human preference data filtered?} 

In~\cite{bai2022training},  the authors mention: ``\emph{We only include comparisons [...] if crowdworkers expressed a preference stronger than the weakest available. [...] we will not otherwise use this preference-strength information; we treat all comparisons in our dataset as binary and of equal weight ([...] we do not include ties).}''. From this, we imply that preferences are excluded if the preferred option is not considerably better than the alternative (rejected). 

\textbf{How are the collected human preference data used to fine-tune (align) the models?} 

In~\cite{bai2022training},  the authors mention that they: `\emph{[...] train a [Reward Model] to assign a higher score to the ‘better’ item in each comparison. [...] each comparison consists of a prompt followed by a pair of model-generated responses, with a [Reward Model] score evaluated at the end of each response.}''. Remember that they collected dialogues, so they use comparisons at the end of its stage of the dialogue. All reward models have been initially trained: ``\emph{on a mixture of comparison data made from StackExchange, Reddit, and Wikipedia.}'' using the communities' up-voting system as a preference signal. 

\textbf{Are the collected human preference data being published?} 

Anthropic has released a dataset, dubbed HH-RLHF,\footnote{\url{https://huggingface.co/datasets/Anthropic/hh-rlhf}}, as part of the work of ~\cite{bai2022training}. The dataset comprises human preference data related to helpfulness and harmlessness, including approx. 160K binary (accepted, rejected) comparisons of dialogue snapshots.

\subsubsection{Google's Gemini}
\label{sec:gemini}

\textbf{What are the selected objectives that developed models should align with?} 

In \cite{team2023gemini}, describing the development of Gemini 1, the authors mention: ``\emph{The post-training recipes are carefully designed to balance multiple objectives, including creativity, factuality, safety and more [...] a particular focus on safety}''. We can draw analogies between factuality and truthfulness, and safety and harmlessness, with creativity understood as part of helpfulness, based on the reported evaluations. Concerning safety, the authors elaborate: ``\emph{safety policies [...] include generation of child sexual abuse and exploitation content, hate speech, harassment, dangerous content such as guidance on how to make weapons, and malicious content. We also aim to reduce bias in our models [...] providing content that reflects our global user base.}'' with the latter part posing a cultural diversity angle. Concerning truthfulness, the authors mention: ``\emph{prioritize providing neutral answers grounded in authoritative, consensus facts, or providing multiple perspectives where consensus doesn’t exist.}'' as fact-based when possible, and nuanced in the lack of consensus. On the contrary, there is no further information on how creativity is operationalized. For Gemini 1.5~\citep{team2024gemini15}, the authors mention: ``\emph{In addition to the models being safe, we also want our models to be helpful and not make assumptions about user intent.}'', while they also mention ``\emph{[We] want Gemini to uphold from a safety, security, and responsibility perspective.}''. Further on, the authors mention: ``\emph{Gemini also has guidelines designed to prevent misinformation or biased content. The model should also [...] prioritize content grounded in authoritative consensus and answers that are neutral (including political neutrality).}''.

\textbf{Who decided on the appropriateness of the selected objectives?} 

For Gemini 1.5~\citep{team2024gemini15}, the authors mention the Google DeepMind Responsible Development and Innovation team, alongside the Google DeepMind Responsibility and Safety Council, which: ``\emph{draw from various sources in producing impact assessments, including a wide range of literature, external expertise, and our in-house ethics and safety research.}'', but in the context of review and evaluation, and not straightforwardly connected to any process related to the selection of objectives (value setting).

\textbf{How are the objectives prioritized when they conflict?} 

In  \cite{team2023gemini}, safety is posed as the main focus, as shown earlier in the definition of objectives mentioning: ``\emph{a particular focus on safety}''; however, there is no information on how this focus is enforced by any means, e.g., guidelines for labelers, post-hoc methods. 

\newpage

\textbf{How is the data annotation process orchestrated?} 

In \cite{team2023gemini}, the authors mention: ``\emph{human raters provide feedback such as relative preferences over candidate responses and feedback regarding individual responses to a given prompt.}''. Without further context available, we imply a very similar ranking setup to \cite{ouyang2022training,bai2022training} and the collection of metadata per individual response. They also mention: ``\emph{In contrast to generic [...] catering to all types of user queries, our safety mitigation is more focused on adversarial, or “harm-inducing” queries [...] likely to produce harmful responses.}'' which sounds similar to the red-teaming efforts of \cite{bai2022training}. For Gemini 1.5~\citep{team2024gemini15}, the authors also briefly mention the collection of comparison data.

\textbf{What is the process of hiring, selecting, and preparing annotators?} 

In \cite{team2023gemini}, the authors mention: ``\emph{For certain data creation and evaluation initiatives, we consider diversity across gender presentation, age, and racial and ethnic diversity.}'', although it is not clear if these criteria are employed when collecting preference data. Furthermore, there is no information concerning training and/or screening labelers, rather than broad statements such as: ``\emph{all data collected meets Google DeepMind’s best practices on data enrichment, developed based on the Partnership on AI’s Responsible Sourcing of Data Enrichment Services.}''.

\textbf{What is the process for selecting the prompts?} 

For Gemini 1.5~\citep{team2024gemini15}, the authors vaguely mention: ``\emph{For both RM and RL [...] we source prompts [...] striving for coverage of safety policies and use cases.}''. 

\textbf{What is the volume of the collected human preference data?} N/A

\textbf{Are the collected human preference data filtered?} N/A

\textbf{How are the collected human preference data used to fine-tune (align) the models?} 

\cite{team2023gemini} mentions: ``\emph{We use this data to train RMs to output rewards that align with human preferences as closely as possible.}''. Similarly, for Gemini 1.5~\citep{team2024gemini15}, the authors mention: ``\emph{Preference data is then amortized in our Reward Model.}''.

\textbf{Are the collected human preference data being published?} 

There are no publicly available datasets related to Gemini initiatives.

\subsubsection{Meta's Llama}

\underline{Note:} The report for the first Llama model~\citep{touvron2023llama1} is excluded, since it does not involve alignment from human feedback.

\textbf{What are the selected objectives that developed models should align with?} 

For Llama 2~\citep{touvron2023llama2}, the authors mention: ``\emph{For our collection of preference annotations, we focus on helpfulness and safety. Helpfulness refers to how well [...] responses fulfill users’ requests and provide requested information; safety refers to whether [...] responses are unsafe}''. Further on, they mention: ``\emph{We asked the annotators to write responses that are informative, truthful, relevant, clear and harmless.}'', which also implies truthfulness. For Llama 3~\citep{llama3}, the authors do not state specific objectives, and the content is overly focused on instruction-following (helpfulness); nonetheless there is a special notice on factuality and avoidance of hallucinations: ``\emph{post-training should align the model to “know what it knows” rather than add knowledge}'', while also safety is mentioned as part of the alignment process.  For Llama 4~\citep{llama4},  the model card mentions: ``\emph{Our approach is to build the most helpful models [...], by aligning our model’s safety for a standard set of risks.}'', while they later mention a set of critical risks including CBRN (Chemical, Biological, Radiological, and Nuclear), child safety, and cybersecurity considerations.

\textbf{Who decided on the appropriateness of the selected objectives?} N/A

\textbf{How are the objectives prioritized when they conflict?} 

In \cite{touvron2023llama2}, the authors mention: \emph{``[We] asked the annotators to prioritize harmlessness over informativeness and helpfulness in cases of prompts that could lead the responses to be problematic''}, when it comes to data collected for SFT. For Llama 3~\citep{llama3}, the authors state: ``\emph{We conduct multiple experiments to determine the optimal ratio of adversarial, borderline, and helpfulness examples, aiming to optimize the trade-off between [response refusals and violations of safety policies]}'', while they also mention: ``\emph{[We] evaluate model performance on helpfulness [...] to ensure that safety improvements do not compromise overall helpfulness.}''.

\textbf{How is the data annotation process orchestrated?} 

In \cite{touvron2023llama2}, the authors mention: ``\emph{we collect human preferences data for reward modeling. We chose a binary comparison protocol over other schemes, mainly because it enables us to maximize the diversity of collected prompts.}'', in more detail: ``\emph{We ask annotators to first write a prompt, then choose between two sampled model responses [...] we also ask annotators to label the degree to which they prefer their chosen response over the alternative: [...] significantly better, better, slightly better, or negligibly better/unsure.''}. The authors collect annotations for helpfulness and safety separately, similar to~\cite{bai2022training}: ``\emph{Separating the two allows us to apply specific guidelines to each and better guide annotators; for example, our safety annotations provide instructions to focus on adversarial prompts [...]}''. They also collect a safety label indicating if one, both, or no response is safe. For Llama 3~\citep{llama3}, the authors mention: ``\emph{We also incorporate an editing step [...] Annotators edit the chosen response directly or prompt the model with feedback to refine its response. Consequently, a portion of our preference data has three responses ranked (edited \textgreater chosen \textgreater rejected).}''. For Llama 4~\citep{llama4},  the model card only vaguely mentions: ``\emph{We employ a multi-faceted approach to data collection, combining human-generated data from our vendors with synthetic data to mitigate potential safety risks.}''. 

\textbf{What is the process of hiring, selecting, and preparing annotators?} 

In \cite{touvron2023llama2}, the authors mention that they screen data labelers via: ``\emph{a multi-step assessment process where we tested their understanding of our guidelines, the alignment with our quality assessment criteria, the alignment with our sensitive topics guidelines, and their reading and writing skills.}''. They further describe tests related to grammar, reading comprehension, and writing style, answering questions on sensitive topics, grading prompt-response pairs, and writing responses for given prompts. All these are related to SFT, and we do not know if they apply to labelers of preference data. There is no information related to the number of annotators or demographics. There is no related information in follow-up work.

\textbf{What is the process for selecting the prompts?} 

In \cite{touvron2023llama2}, the authors mention: ``\emph{We ask annotators to first write a prompt}'', i.e., the prompts are collected on-the-fly, similar to~\cite{bai2022training}.  For Llama 4~\citep{llama4},  the model card vaguely mentions: ``\emph{We’ve developed [...] (LLM)-based classifiers that enable us to thoughtfully select high-quality prompts and responses}''.

\textbf{What is the volume of the collected human preference data?} 

In \cite{touvron2023llama2}, the authors mention the collection of approx. 1.4M comparisons for both safety and helpfulness. For Llama 3~\citep{llama3}, the authors state ``\emph{[We] improved both the quantity and quality of the data we use for pre-training and post-training.}''. Further on, the authors state: ``\emph{As [we] progress, we develop stronger [models] that we use to collect larger datasets}'', to the point where: ``\emph{most of our training data is model-generated}''. 

\textbf{Are the collected human preference data filtered?} 

In \cite{touvron2023llama2}, the authors mention: ``\emph{We have implemented a quality assurance process to ensure we only use high-quality annotations for training the model. [...] a team of highly skilled content managers manually reviewed the annotations and approved the ones that would be used.}'', which seems to refer only to the collected data for SFT. For Llama 3~\citep{llama3}, the authors mention: ``\emph{Following Llama 2, we use all of our preference data for reward modeling after filtering out samples with similar responses.}''. Later on, they describe several steps of cleaning and quality control, which include: the removal of undesirable data, i.e., including many emojis, or overly apologetic responses,  the pruning of low-quality and low-difficulty pairs, and semantic deduplication. It is unclear if those practices apply to the preference data meant for reward modeling or only for SFT. For Llama 4~\citep{llama4},  the model card vaguely mentions: ``\emph{We’ve developed [...] (LLM)-based classifiers that enable us to thoughtfully select high-quality prompts and responses, enhancing data quality control.}''.

\textbf{How are the collected human preference data used to fine-tune (align) the models?}

In \cite{touvron2023llama2}, the authors train reward models where they: ``\emph{[...] convert our collected pairwise human preference data into a binary ranking label format [...] and enforce the chosen response to have a higher score than its counterpart.}''. The additional labeling (4 degrees of preference) collected was used ``\emph{to explicitly teach the reward model to assign more discrepant scores to the generations that have more differences. [...] we further add a margin component in the loss}''. The authors used a mixture of their own collected data (see above), alongside publicly available data such as Anthropic's HH-RLHF~\citep{bai2022training} data, data released from OpenAI~\citep{Stiennon2020,nakano2021webgpt}, Stanford's SHP dataset~\citep{ethayarajh22a}, and data from StackExchange. The authors mention: ``\emph{Initially, open-sourced datasets were used to bootstrap our reward models while we were in the process of collecting preference annotation data.}''. There is no related information in follow-up work.

\textbf{Are the collected human preference data being published?} 

There are no publicly available datasets related to Meta's Llama initiatives.

\subsubsection{Google's Gemma}

\textbf{What are the selected objectives that developed models should align with?} 

In \cite{team2024gemma1}, the authors mention in the introduction: ``\emph{we release [...] checkpoints fine-tuned for dialogue, instruction-following, helpfulness, and safety.}''. Later on, the authors state: ``\emph{Different prompt sets are constructed to highlight specific capabilities, such as instruction following, factuality, creativity, and safety.}'' related to SFT. The authors state several times that they follow the work of~\cite{team2023gemini} (discussed in Section~\ref{sec:gemini}). For Gemma 2~\citep{team2024gemma2}, the authors state: ``\emph{The final data mixtures and post-training recipe [...] were chosen on the basis of improving helpfulness while minimizing model harms related to safety and hallucinations.''}. They also vaguely mention: ``\emph{The new reward model is also oriented more towards conversational capabilities, specifically multi-turn.}''. For both Gemma 2 and 3~\cite{team2024gemma1}, the authors include a more elaborate section, which better describes the focus of harmlessness and safety policies mentioning the areas of: child abuse, Personally Identifiable Information (PII), hate speech and harassment, dangerous or malicious content, sexually explicit content, and medical advice against scientific consensus.

\textbf{Who decided on the appropriateness of the selected objectives?} N/A

\textbf{How are the objectives prioritized when they conflict?} N/A

\textbf{How is the data annotation process orchestrated?} 

In \cite{team2024gemma1}, the authors briefly mention: ``\emph{We collected pairs of preferences from human raters}''.

\textbf{What is the process of hiring, selecting, and preparing annotators?} N/A

\textbf{What is the process for selecting the prompts?} N/A

\textbf{Are the collected human preference data filtered?} N/A

\textbf{What is the volume of the collected human preference data?} N/A

\textbf{How are the collected human preference data used to fine-tune (align) the models?} 

In \cite{team2024gemma1}, the authors mention: ``\emph{We collected pairs of preferences [..] and trained a reward function under the Bradley-Terry model~\citep{bradley1952rank}, similarly to Gemini.}'' 

\textbf{Are the collected human preference data being published?} 

There are no publicly available datasets related to Gemma initiatives.

\underline{Note:} The information we get from the documentation of Gemma models is very limited, compared to the rest of the development initiatives. We can account for this limitation to the fact that Google is the only organization running two initiatives, a closed-source/weight and an open-weight one, namely the Gemini and Gemma models. In other words, the work of \cite{team2023gemini} can be considered as the foundation of Gemma models, as well.

\subsubsection{Alibaba's Qwen}

\underline{Note:} The report for the Qwen 1 models~\citep{qwen1technicalreport} is excluded, since it does not involve alignment from human feedback.

\vspace{2mm}

\textbf{What are the selected objectives that developed models should align with?} 

In \cite{qwen2technicalreport} describing the development of Qwen 2, the authors mention: ``\emph{we engage in a post-training phase [...] it ensures that the generation from the models is in harmony with human values, making it helpful, honest, and harmless}''. For Qwen 2.5~\citep{qwen25technicalreport}, the authors provide more elaborate descriptions of the objectives: ``\emph{we adhere to a set of carefully defined labeling criteria [...] that the responses generated by the model are [...] aligned with ethical and user-centric standards~\cite{wang2024secrets}. The specific guidelines for data labeling are [...]: Truthfulness [...], Helpfulness [...], Conciseness [...], Relevance [...], Harmlessness [...], Debiasing [...]}'', where each aspect is described with a brief paragraph. For Qwen 3~\citep{qwen3technicalreport}, the authors mention vaguely: ``\emph{preference alignment focuses on improving the model’s helpfulness, engagement, and style, ultimately delivering a more natural and satisfying user experience.}''.

\textbf{Who decided on the appropriateness of the selected objectives?} N/A

\textbf{How are the objectives prioritized when they conflict?} N/A 

\textbf{How is the data annotation process orchestrated?} 

In \cite{qwen2technicalreport}, the authors mention: ``\emph{Multiple responses to an instruction are obtained [...]. Annotators rank these responses based on their preferences [...], yielding both demonstration and preference data.}''

\textbf{What is the process of hiring, selecting, and preparing annotators?} N/A

\textbf{What is the process for selecting the prompts (used to generate model responses and gather user preferences)?} 

In \cite{qwen2technicalreport}, the authors describe a protocol for selecting prompts, specifically: ``\emph{The process initiates with the application of [...] an open-set fine-grained tagger, to extract the underlying ontology from a large-scale instruction dataset. [...] Each instruction, with tags annotated, is evaluated for tag diversity, semantic richness, complexity, and intent completeness. Based on these criteria, we select a set of representative instructions [...] To enrich the instruction dataset, a self-evolution strategy~\citep{zhao2024tree} is employed, [...], thereby increasing their complexity and ensuring a diverse range of difficulty levels within the dataset.}''. Although it is unclear if this protocol is employed to collect prompts (instructions) across all post-training stages, or not.

\textbf{What is the volume of the collected human preference data?} N/A

\textbf{Are the collected human preference data filtered?} N/A

\textbf{How are the collected human preference data used to fine-tune (align) the models?}  

In \cite{qwen2technicalreport}, the authors mention: ``\emph{Our training regime for RLHF comprises two sequential stages: offline and online training. In the offline training stage, we use a pre-compiled preference dataset [...] with Direct Preference Optimization (DPO, \citet{rafailov2023direct}). In the online training stage, the model iteratively refines its performance [...] leveraging reward models for immediate feedback. [...] the reward model selects the most and the least preferred responses, forming preference pairs that are used for DPO in each episode.}''

\textbf{Are the collected human preference data being published?} 

There are no publicly available datasets related to Qwen initiatives.

\subsection{Overall Findings}
\label{sec:overall_findings}

Most of the information comes from the early initiatives~\citep{ouyang2022training, bai2022training, touvron2023llama2}, as the depth and quality of information we get from follow-up work, i.e., related to more recent versions of the models, is very limited, since the newer documentation (mostly tech reports or model/system cards rather than academic-standard articles) are heavily focused on presenting benchmarking results, and new capabilities, e.g., processing and generating images, and other modalities, or using tools, etc.

\begin{figure*}[t]
    \centering
    \includegraphics[width=\textwidth]{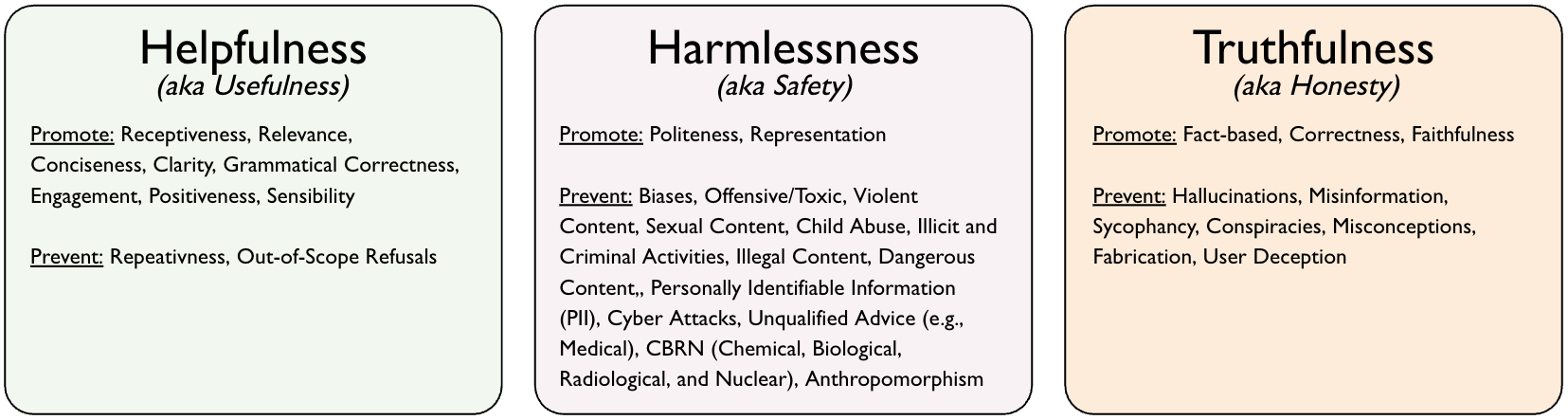}
    \caption{The three pillars of alignment in the examined work: \emph{helpfulness}, \emph{harmlessness}, \emph{truthfulness}, and the core underlying aspects they aim to promote or prevent.}
    \label{fig:objectives}
\end{figure*}

\paragraph{Objectives} As we observe among all initiatives (Section~\ref{sec:per_initiative}), there are three pillars of AI Alignment in the examined body of work: namely \emph{helpfulness}, \emph{harmlessness}, and \emph{truthfulness} (Figure~\ref{fig:objectives}), following the foundational HHH framework of~\cite{askell2021general}. Since these objectives are quite broad, we review them in more detail based on the examined work:

(a) \emph{Helpfulness} aims to help users solve tasks effectively and efficiently. It promotes model responses that are relevant to the instruction, concise (comprehensive), and clear (non-ambiguous). Other secondarily related criteria that the model promotes, mentioned in the examined work, are positiveness and user engagement. The model should avoid repetitiveness, and out-of-scope refusals, i.e., refuse to answer when the model is capable of answering, and the response does not violate safety policies. 

(b) \emph{Harmlessness} is concerned with avoiding inflicting harm to the users and promoting safety at large, i.e., not weaponizing the user to harm others or themselves.\footnote{This can be understood as a framework related to `The Harm Principle'~\citep{mill1859liberty}, a central concept to liberalism, in an expanded form since harmlessness also accounts for self-harm in this case.} In general, it is mostly a `preventive' moderational objective. It mainly concerns the avoidance of biases and stereotypes, the use of offensive, derogatory, or toxic language, and the presentation of sexually explicit or violent content. The models should avoid sharing Personally Identifiable Information (PII), as also avoid sharing unqualified advice, e.g., medical advice, against scientific consensus. There is also the aspect of child safety and the prevention of child abuse. More recently, safety efforts also account for the so-called CBRN (Chemical, Biological, Radiological, and Nuclear) threats or hazards. With the models trained to have agentic capabilities and being able to write or execute code, and use tools, there are also efforts against illicit use of such capabilities, mostly related to cyberattacks. Another element has to do with anthropomorphism, i.e., the model portraying itself as being a human actor, having agency and feelings, etc. The only aspects of harmlessness that can be considered as being `promoted' rather than `preventive' are politeness and, in a few cases, the representation of different identities. 

(c) \emph{Truthfulness} aims to promote the sharing of fact-based knowledge and information, and correctness, while preventing hallucinations (confabulations), i.e., the system making things up, alongside the spread of misinformation, conspiracy theories, and misconceptions. Extremely relevant to the objective of truthfulness, and its relation to honesty--which most times is used interchangeably--is the undesired phenomenon of \emph{sycophancy}, the obsequious behavior towards user preferences to reinforce engagement. Sycophancy can also be understood as related to user deception.

The three pillars (helpfulness, harmlessness, and truthfulness), originally proposed by~\cite{askell2021general}, have not been contested or considerably altered at any point in any work. Although the scope of the objectives has become broader with additional considerations, e.g., the focus on preventing CBRN (Chemical, Biological, Radiological, and Nuclear) threats or hazards, and cyber-attacks, both related to harmlessness (safety), and the mitigation of sycophancy and deception related to honesty, and indirectly to harmlessness.

As a noteworthy exception, we see the work of Anthropic's Claude, where Claude’s constitution~\citep{bai2022constitutional} includes statements echoing ones from the UN Declaration of Human Rights, among other statements that encourage the consideration of non-Western perspectives, and ethical and moral awareness; some of those not directly connected to any of the main pillars. Similarly, we cannot establish a straightforward connection between creativity (discussed in Google's initiatives) and helpfulness. 

\paragraph{Objectives' Authority} Across all projects, the objectives are most likely selected by the development team, as a proxy of the corporation (employer) itself. In half of the cases, there is no verbatim statement on this aspect. This could be considered problematic because it signifies a lack of accountability and a point of potential confusion. Early initiatives~\citep{ouyang2022training, bai2022training} have been open about this design choice and mention related concerns (excessive authority, concentration around AI developers' concerns, lack of broader considerations), while follow-up work does not reiterate concerns over this topic. 

Since all initiatives follow the framework of~\cite{askell2021general}, in one way or another, we can also consider \citeauthor{askell2021general} -as a group of individuals or as a proxy of Anthropic as an organization- as a highly influential `undeclared authority' and understand this design choice by the examined work as simply following the literature (prior art and best practices). In many cases, there is also an effort to offload part of the authority to the data labelers, since they are the ones making the final choices (preferences); in other words, their data collectively is seen as the materialization of the alignment objectives. While acknowledging that data labelers indeed have a substantial level of discretion in the process, their behavior is heavily ``groomed'' (biased) by the initiative guidelines and onboarding.

In most cases, there is no claim that the objectives represent `human values'. There are a few exceptions in the work of (a) Anthropic's Claude, where the inclusion of the UN Declaration of Human Rights as part of the Claude’s constitution is accounted as a basis of declaring that their models align with ``human values''~\citep{claude2}.\footnote{Similar framing in \url{https://www.anthropic.com/news/claudes-constitution}: ``\emph{[2] The UN declaration of Human Rights, having been drafted by representatives with different legal and cultural backgrounds and ratified (at least in part) by all 193 member states of the UN, seemed one of the most representative sources of human values we could find.}''}, and (b) Alibaba's Qwen team~\citep{qwen2technicalreport} state: ``\emph{[post-training] ensures [...] harmony with human values}''.\footnote{In~\cite{qwen3technicalreport}, there is another related broader unfounded statement: ``\emph{the pursuit of artificial general intelligence (AGI) or artificial super intelligence (ASI) has long been a goal for humanity.}''.}

\paragraph{Prioritization of Objectives} In a few cases, we see that data labelers are instructed to prioritize harmlessness over helpfulness in early initiatives, while other work tries to balance out their trade-offs post-hoc, i.e., optimize data mixture, etc., to minimize model response refusals in parallel to minimizing the violation of safety policies. In general, data labelers have substantial discretion when performing their tasks, although trained and screened in most cases. Overall, harmlessness (safety) seems to be prioritized over helpfulness across all models, through alignment or other means (as a system overall), but we do not find any elaborate discussion on the overall topic.

\paragraph{Preparation of Annotators} In a few early initiatives~\citep{ouyang2022training, bai2022constitutional, touvron2023llama2}, we have a considerable level of information related to the hiring, training, and screening of data labelers. In such cases, candidates pass a screening phase where they are assessed based on reading comprehension, understanding of guidelines, and their agreement with the development team's ratings over test cases. After this screening phase, the data labelers have substantial discretion when performing their tasks. 

\cite{bai2022training} is the only work acknowledging loose selection criteria (no official screening) in the hope of data diversity and the ‘wisdom of the crowd’. \cite{ouyang2022training} is the only work sharing the actual guidelines as a document, while in the work of \cite{bai2022training}, the authors share excerpts of the guidelines and screenshots of the annotator instructions.

A few early initiatives~\citep{ouyang2022training, bai2022constitutional} also share more details on hiring processes and demographics of data labelers. The hiring usually occurs via well-known vendors (e.g., MTurk, UpWork), and the population is 30-40 individuals. The population is mostly equally split between males/females, with a few individuals identifying as LGBTQ+. The vast majority of annotators are under 35 years old and highly educated (graduate, post-graduate). In one case, the annotators were US-based, while in another work, there was a majority of people residing in Southeast Asia and US. We have no information related to recent projects, coming from the examined work.

\paragraph{Annotation Approach} All initiatives follow the same paradigm, collecting human-written demonstrations and preference (comparison) data, where given a prompt (instruction), two or more alternative model-generated responses are presented and the labelers have to compare and rank them. An advanced solution is presented in~\cite{bai2022training,llama3}, where two alternative model-generated responses are compared, but the annotators also label the degree of preference, i.e., (significantly/slightly/negligibly) better.

In a few cases, development teams collect additional metadata as part of the process, usually related to safety~\citep{ouyang2022training, touvron2023llama1}, e.g., if a model-generated response is generally unsafe, or fine-grained, i.e., sexually explicit, gives harmful advice, etc. As an additional step, ~\cite{bai2022training,llama3} asked annotators to optionally rewrite (refine) the approved suggested model-generated response to provide a better human-curated alternative. 

\paragraph{Selection of Prompts} In most early initiatives~\citep{bai2022training, touvron2023llama2}, there is no selection of the prompts, since in most cases the data annotators evaluating model responses (output) were also the authors of the prompts (input), except \cite{ouyang2022training}, where the authors relied on prompts' availability via their platform (OpenAI Playground). \cite{qwen2technicalreport} and \cite{llama4} mention the selection of prompts from large-scale instruction datasets using data filtering for de-duplication, and the promotion of instruction diversity or the enhancement of those with older models, but it is not clear if this practice was employed for comparison data, or merely the generation of human responses for SFT. 

The process of prompt selection is overall under-documented, considering that it has a substantial impact on the process from many different perspectives, e.g., coverage, effort, etc.,~\citep{boubdir2023promptsmakedifferencedata}.

\paragraph{Data Volume} In early initiatives~\citep{ouyang2022training, bai2022training}, the collected data are in the scale of 50-200K pairs, while later work, such as~\cite{touvron2023llama2} and \cite{qwen25technicalreport}, scales to millions of pairs. In many cases, the development teams use other publicly available alignment-related datasets, such as the Stack Exchange dataset~\citep{h4stackexchange}, comprising millions of comparisons between answers from the online platform based on the up-voting system. Such data can be used in a mixture with gold-standard labeled datasets, or as an initial batch to warm-start the reward models. 

LLM development increasingly relies on synthetic (model-generated) data for alignment purposes, and not only, with the majority of the data being synthetic at this point~\citep{llama3,qwen25technicalreport}.

\paragraph{Data Filtering} Data filtering regarding alignment-focused data is very under-specified alongside the examined body of work. Based on the limited information we have, the main approaches are: (a) de-duplication of samples based on similar prompts~\citep{ouyang2022training, llama3}; (b) exclusion of preferences in the lowest ranking tier and ties~\citep{bai2022training}; and (c) quality assurance reviewing (rejection) of samples, related to data meant for SFT~\citep{touvron2023llama2,llama3}, rather than data meant for alignment. 

\paragraph{Use of Data}

In early work~\citep{ouyang2022training,bai2022training, touvron2023llama2, achiam2023gpt}, RLHF is operationalized as a two-step pipelined process. Initially, the collected data is used to train reward models in a binarized format following the Bradley-Terry model~\citep{bradley1952rank}.\footnote{In most cases, prior instruction-tuned generative LLMs are fine-tuned as reward models.} Then, reinforcement learning based on Proximal Policy Optimization (PPO)~\citep{christiano-etal-2017-drl} is used to fine-tune (align) the instruction-tuned LLMs, or earlier versions of aligned models, to maximize the learned reward while staying close to a reference policy (via KL-divergence constraints). Given recent advances, many later initiatives~\citep{llama3, qwen2technicalreport} follow a simpler process, namely Direct Preference Optimization (DPO)~\citep{rafailov2023direct}, that eliminates the need explicit reward modeling, i.e., does not involve the training of reward models; instead the instruction-tuned LLMs implicitly encode the reward to align with preferences using the preference data. Training reward models is still a possibility as a means of filtering data or bootstrapping more training data, and more.

When comparisons across more than two responses are collected, the aim is to accelerate the annotation process, i.e., collect multiple binary comparisons in a single step, rather than utilizing the multi-response ranking information, except for \cite{touvron2023llama2}, which utilizes the information as an additional margin component. In the case of ranking ties, the pair is excluded, similarly to preferences that are ranked in the lowest tier. 

Most recent work~\citep{claude4,openai2025gpt5} does not mention how human-collected data are used, if human-collected data are used at all. 

\paragraph{Publication of Data} The work of~\cite{bai2022training} is the only initiative that publicly shares data. All other work abstains, while two very early -prior to InstuctGPT 3.5~\citep{ouyang2022training}- OpenAI initiatives have shared data~\citep{Stiennon2020, nakano2021webgpt}. This is very much expected, coming from an understanding of data as valuable private property, under fierce corporate competition, rather than artifacts that are meant to be shared and examined based on open-science standards.

Although we lack publicly released datasets from the examined LLM development initiatives, there are available datasets released, such as Stanford's SHP dataset~\citep{ethayarajh22a}, and NVIDIA's HelpSteer datasets~\citep{wang2024helpsteer2,wang2025helpsteer3preferenceopenhumanannotatedpreference}; nonetheless, their similarity with the ones developed in the examined work is unclear.

\begin{table*}[t]
    \centering
    \resizebox{\textwidth}{!}{
    \begin{tabular}{cc|c|cccccccccc}
         \multicolumn{2}{c|}{\textbf{Initiative / Project}} & \multirow{2}{*}{\textbf{V.}} & \multicolumn{10}{c}{\textbf{Questions}} \\
          \textbf{Company} & \textbf{Name} & & DEF. & AUTH & VS. & PLAN & HUM & PROMPT & SIZE & FILTER & USE & PUB. \\
         \midrule
         \multirow{4}{*}{OpenAI} & \multirow{4}{*}{GPT} & 3.5 & \cellcolor{light_green} & \cellcolor{light_green} & \cellcolor{mid_yellow} & \cellcolor{best_green} & \cellcolor{best_green} & \cellcolor{mid_yellow} & \cellcolor{best_green} & \cellcolor{light_green} & \cellcolor{best_green} & \cellcolor{bad_red} \\
          & & 4 & \cellcolor{mid_yellow} & \cellcolor{mid_yellow}  & \cellcolor{bad_red}  & \cellcolor{mid_yellow} & \cellcolor{bad_red} & \cellcolor{bad_red} & \cellcolor{bad_red} & \cellcolor{bad_red} & \cellcolor{bad_red} & \cellcolor{bad_red} \\
          & & 4.5  & \cellcolor{bad_orange} & \cellcolor{bad_red}  & \cellcolor{bad_red} & \cellcolor{bad_red} & \cellcolor{bad_red} & \cellcolor{bad_red} & \cellcolor{bad_red} & \cellcolor{bad_red} & \cellcolor{bad_red} & \cellcolor{bad_red}\\
          & & 5  & \cellcolor{bad_orange} & \cellcolor{bad_red}  & \cellcolor{bad_red} & \cellcolor{bad_red} & \cellcolor{bad_red} & \cellcolor{bad_red} & \cellcolor{bad_red} & \cellcolor{bad_red} & \cellcolor{bad_red} & \cellcolor{bad_red}\\
          \midrule
          \multirow{4}{*}{Anthropic} & \multirow{4}{*}{Claude} & 1 & \cellcolor{mid_yellow} & \cellcolor{light_green} & \cellcolor{bad_orange} & \cellcolor{best_green} & \cellcolor{best_green} & \cellcolor{mid_yellow} & \cellcolor{light_green} & \cellcolor{mid_yellow} & \cellcolor{best_green} & \cellcolor{light_green} \\
          & & 2 & \cellcolor{mid_yellow}  & \cellcolor{bad_red}  & \cellcolor{bad_red}  & \cellcolor{bad_red} & \cellcolor{bad_red} & \cellcolor{bad_red}  & \cellcolor{bad_red}  & \cellcolor{bad_red}  & \cellcolor{bad_red} & \cellcolor{bad_red} \\
          & & 3.x & \cellcolor{bad_orange}  & \cellcolor{bad_orange}  & \cellcolor{bad_red}  & \cellcolor{bad_red} & \cellcolor{bad_red} & \cellcolor{bad_red}  & \cellcolor{bad_red}  & \cellcolor{bad_red}  & \cellcolor{bad_red} & \cellcolor{bad_red} \\
          & & 4 & \cellcolor{bad_orange} & \cellcolor{bad_red} & \cellcolor{bad_red}  & \cellcolor{bad_red} & \cellcolor{bad_red} & \cellcolor{bad_red}  & \cellcolor{bad_red}  & \cellcolor{bad_red}  & \cellcolor{bad_red}  & \cellcolor{bad_red}\\
          \midrule
          \multirow{4}{*}{Google} & \multirow{4}{*}{Gemini}  & 1 &  \cellcolor{mid_yellow}  & \cellcolor{bad_red}  & \cellcolor{bad_orange}   & \cellcolor{mid_yellow} & \cellcolor{bad_orange} & \cellcolor{bad_red}  & \cellcolor{bad_red}  & \cellcolor{mid_yellow}  & \cellcolor{bad_red} & \cellcolor{bad_red}\\
          & & 1.5 & \cellcolor{light_green} & \cellcolor{mid_yellow} & \cellcolor{bad_red} & \cellcolor{mid_yellow} & \cellcolor{bad_red} & \cellcolor{bad_orange} & \cellcolor{bad_red} & \cellcolor{bad_red} & \cellcolor{mid_yellow} & \cellcolor{bad_red} \\
          & & 2.5  &  \cellcolor{bad_orange}  & \cellcolor{bad_red}  & \cellcolor{bad_red}  & \cellcolor{bad_red} & \cellcolor{bad_red} & \cellcolor{bad_red}  & \cellcolor{bad_red}  & \cellcolor{bad_red}  & \cellcolor{bad_red} & \cellcolor{bad_red}\\
          \midrule
          \multirow{4}{*}{Meta} & \multirow{4}{*}{Llama} & 2 &\cellcolor{light_green}  & \cellcolor{bad_red}  & \cellcolor{light_green}  & \cellcolor{best_green} & \cellcolor{light_green} & \cellcolor{best_green} & \cellcolor{light_green}  & \cellcolor{mid_yellow} & \cellcolor{light_green} & \cellcolor{bad_red}\\
          & & 3.x & \cellcolor{bad_orange}  & \cellcolor{bad_red}  & \cellcolor{mid_yellow}   & \cellcolor{light_green}  & \cellcolor{bad_red} & \cellcolor{bad_red} & \cellcolor{best_green} & \cellcolor{mid_yellow} & \cellcolor{best_green} & \cellcolor{bad_red} \\
          & & 4 & \cellcolor{mid_yellow} & \cellcolor{bad_red} & \cellcolor{bad_red} & \cellcolor{bad_orange} & \cellcolor{bad_red} & \cellcolor{bad_red} & \cellcolor{bad_orange} & \cellcolor{bad_red} & \cellcolor{bad_red} & \cellcolor{bad_red} \\
          \midrule
          \multirow{4}{*}{Google} & \multirow{4}{*}{Gemma} & 1 & \cellcolor{bad_orange}  & \cellcolor{bad_red} & \cellcolor{bad_red} & \cellcolor{bad_orange}  & \cellcolor{bad_red}  & \cellcolor{bad_red}  & \cellcolor{bad_red}  & \cellcolor{mid_yellow}  & \cellcolor{bad_red}  & \cellcolor{bad_red}\\
          & & 2 & \cellcolor{mid_yellow}  & \cellcolor{bad_red} & \cellcolor{bad_red}  & \cellcolor{bad_red}  & \cellcolor{bad_red}  & \cellcolor{bad_red}  & \cellcolor{bad_red}  & \cellcolor{bad_red}  & \cellcolor{bad_red}  & \cellcolor{bad_red} \\
          & & 3 & \cellcolor{mid_yellow}  & \cellcolor{bad_red} & \cellcolor{bad_red}  & \cellcolor{bad_red}  & \cellcolor{bad_red}  & \cellcolor{bad_red}  & \cellcolor{bad_red}  & \cellcolor{bad_red}  & \cellcolor{bad_red}  & \cellcolor{bad_red}\\
          \midrule
          \multirow{4}{*}{Alibaba} & \multirow{4}{*}{Qwen} & 2 & \cellcolor{mid_yellow}  & \cellcolor{bad_red} & \cellcolor{bad_red} & \cellcolor{mid_yellow}  & \cellcolor{bad_red} & \cellcolor{bad_red} & \cellcolor{bad_red}  & \cellcolor{bad_red}  & \cellcolor{light_green} & \cellcolor{bad_red} \\
          & & 2.5 & \cellcolor{light_green}  & \cellcolor{bad_red} & \cellcolor{bad_red} & \cellcolor{bad_red}  & \cellcolor{bad_red} & \cellcolor{mid_yellow} & \cellcolor{bad_red}  & \cellcolor{bad_red}  & \cellcolor{light_green} & \cellcolor{bad_red} \\
          & & 3 &  \cellcolor{bad_orange} & \cellcolor{bad_red} & \cellcolor{bad_red} & \cellcolor{bad_red}  & \cellcolor{bad_red} & \cellcolor{bad_red} & \cellcolor{bad_red}  & \cellcolor{bad_red}  & \cellcolor{light_green} & \cellcolor{bad_red} \\
    \end{tabular}
    }
    \caption{Heatmap of the quality (depth) of information per question (Section~\ref{sec:questionaire}) across the examined LLM development initiatives as presented in Section~\ref{sec:per_initiative}. There is a keyword per question: Q1 (DEF.), Q2 (AUTH), Q3 (VS.), Q4 (PLAN), Q5 (HUM), Q6 (PROMPT),  Q7 (SIZE), Q8 (FILTER), Q9 (USE), Q10 (PUB).}
    \label{tab:heatmap}
\end{table*}

\subsubsection{Quality of Information}
\label{sec:quality}

To get a broader picture regarding the quality (depth) of information per initiative across time, we grade the information provided by each LLM project per question to get an overall idea. We follow a Likert scale with 5 degrees: (a) 0: There is no information available, (b) 1: The information is limited and very vague, (c) 2: The information is limited and mostly vague, (d) 3: The information is more or less complete with a few vague (unclear) points, (e) 4: The information is complete, meeting open-science standards. In Table~\ref{tab:heatmap}, we see a heatmap, where red color refers to category 0 (worst-case), and dark green to category 4 (best-case). As we observe, only three out of six initiatives (OpenAI's GPT, Anthropic's Claude, and Meta's Llama) have disclosed high-quality information in early projects, while the information sharing related to alignment-focused processes substantially declined or stopped in follow-up cycles, one among the several concerns we discuss in the next section.

\section{Concerns}
\label{sec:concerns}

We discuss a series of concerns related to the overall findings around the operationalization of alignment in LLM development initiatives. To do so, we consider several aspects and present related work that points out open challenges, limitations, and in some cases suggest alternatives on how alignment can be envisioned, designed, and orchestrated from another perspective, in most cases, more participatory (democratic), considerate of social challenges, and closer to open-science principles.

\paragraph{The information stream is draining} As already mentioned, most of the information comes from early initiatives~\citep{ouyang2022training, bai2022training, touvron2023llama2} in 2022-2023, with more recent work sharing limited or no information related to alignment-focused processes. We may hypothesize that this is mainly a byproduct of industrial secrecy, i.e., keeping a competitive edge or non-disclosure for other reasons, nor ignorance for knowledge sharing, but this phenomenon mostly seems connected to three other factors:\footnote{Although such dimensions are not rejected by any means.}

(a) Recent work is heavily concentrated on improving reasoning capabilities and developing new capabilities related to the processing of other modalities rather than text, e.g., images, speech, or efforts on agentic capabilities, i.e., function-calling, use of tools, and multi-agent environments. In other words, the information drain is also tied to a shift in focus, where alignment is only documented in connection to evaluations and red-teaming efforts.

(b) The possibility that there is nothing fundamentally novel and noteworthy regarding alignment-focused processes related to human feedback, which is connected to the aforementioned point (a)--focus swift--. The companies use similar frameworks with some differences, e.g., use of new, larger datasets, consideration of other side aspects related to the main objectives, i.e., extended safety policies, and the use of new optimization algorithms, etc.

(c) The extensive use of practices related to Reinforcement Learning from AI Feedback (RLAIF) practices. We see more and more the use of prominent techniques, such as Constitutional AI~\citep{bai2022constitutional}. Since models became heavily capable and steerable given appropriate prompting, the use of synthetic (model-generated) data is gaining ground over utilizing human-crafted data, to the point where most of the training data is model-generated, as acknowledged in the examined work~\citep{llama3,qwen25technicalreport}. Similarly, the use of automated verifiable objectives related to mathematical and coding reasoning is extensively used to enhance model capabilities. 

While there is active research related to several aspects of alignment of LLMs, for instance, related to \emph{pluralistic alignment}~\citep{sorensen2024roadmap, conitzer2024social}, or considerations on participatory definition of objectives and data curation~\citep{kirk2024prism, huang2024collective, don2025future}, we do not yet see any operationalization of such ideas by the established mainstream development initiatives.

\paragraph{Objectives decided in a vacuum} As we discuss in Section~\ref{sec:overall_findings}, alignment objectives are decided internally without broader consideration that are not related to the productization of LLMs.\footnote{The three pillars (HHH) mainly try to cover the productization of LLMs, as a minimal framework that enables LLMs to be considerably useful and safe} Although most of the initiatives have established alignment and safety teams, there is no yet consideration of broader society-wide consolidation, in the form of liberal society, independent organizations, or state institutions. Early initiatives~\citep{ouyang2022training,bai2022training, achiam2023gpt} were quite open on the challenges and limitations of developed-designed objectives, leading to developers having a disproportionate (excessive) authority on critical work, the heavy concentration of objectives around AI developers' concerns and beliefs, and the lack of broader consensus and legitimization.\footnote{Similar concerns can be found in an article by~\cite{deliberativealignment}: ``\emph{Today tech companies are making these decisions \underline{largely unilaterally} [...]. To improve alignment with group preferences, we need to set up a process that actually includes humanity when aligning our AI systems. If we do nothing, \underline{we are at risk of these value questions being answered by commercial incentives}}''.} A position that later work retracted from, or at least ignores to mention (document) as a minimal demonstration of accountability. This critical issue has raised concerns, as we mainly experience what we can frame as \emph{corporate alignment}~\citep{bommasani-2021,deliberativealignment,GokselMona2025}, i.e., models are aligned with corporate-inspired objectives that do not necessarily align with the priorities and aspirations (interests) of society at large.

As stated above, there is an increasing body of work related to \emph{pluralistic alignment}~\citep{sorensen2024roadmap, conitzer2024social,sorensen2025valueprofilesencodinghuman}, concerned with designing AI systems that align with objectives under a framework that aims to respect and accommodate diverse values, beliefs, and preferences of many different people and groups based on different approaches (overton consensus, distribution-aware, personalization).

Similarly, there are considerations on participatory objectives and data curation~\citep{kirk2024prism, huang2024collective}, such as studies that try to imitate public deliberation and incorporate public input in deciding the desired alignment objectives, instead of solely corporate/developer-defined ones. 

An interesting ``futurist'' framework on aligning  AI systems with society's values was unofficially proposed in an article by~\cite{deliberativealignment}, coined as \emph{simulated deliberative democracy}--that seems influential in the work of~\cite{huang2024collective}--, while a quite similar idea was explored in the work of~\cite{habermasmachine}. In such a framework, humans share their thoughts (values), and deliberation is practically offloaded to AI systems, seemingly based on the premise that humans somehow lack the capacity to deliberate effectively.\footnote{\cite{habermasmachine} state "``\emph{Collective deliberation can be effectively supported by structured events, such as citizens’ assemblies, \underline{but such events are expensive, are difficult to scale}, and can result in voices being heard unequally}'', in other words a huge part of the motivations comes from an understanding of democracy as a technocratic challenge with logistical limitations, and AI comes once more to the rescue.} The latter point can be understood as a manifestation of the deeper crisis of liberal democracy in our days, where AI comes as an external ``hack''. Such a technocratic approach is inevitable when model development, reflection, and policy come from the very same institutions (tech corporations). The recent work of~\cite{don2025future} seems to promote a more promising, less technocratic, human-centered idea of `open human feedback' towards building sustainable systems of open diverse human feedback, developed under open-science principles. Nonetheless, the governance of such a framework, and how it will escape corporate (elite) capture~\citep{taiwo2022elite} is unclear.

\newpage

\paragraph{Objective prioritization is understudied} Given that a certain level of discretion is needed when deciding how to prioritize objectives in practice~\citep{buyl2025ai}, we expect that this challenge would be addressed early on from when data labelers are on-boarded. We find that objective prioritization is understudied, except for broad instructions to data labelers on how to prioritize harmlessness over helpfulness or examining their trade-offs in a post-hoc manner. \cite{buyl2025ai} argue that current alignment approaches allow an excessive, unscrutinized amount of discretion to both human and algorithmic preferences--when models judge content--and call for the development of interpretable and controllable approaches to alignment discretion, taking inspiration from jurisprudence. 

\paragraph{Beyond Helpfulness-Harmlessness-Honesty (HHH)} As we extensively described, the scope of alignment in LLMs' development is limited to the three pillars of \emph{helpfulness}, \emph{harmlessness}, and \emph{truthfulness} (\emph{honesty}) (HHH), with very few exceptions. \cite{askell2021general} who came up with the HHH clearly state: ``\emph{We chose [HHH] as criteria because they are simple and memorable, and seem to capture the majority of what we want from an aligned AI.}''. This communicates how the HHH framework is not meant to be complete or elaborate by any means, solely reflects the authors' understanding, and is used as an experimental ``playground'' for AI Alignment (`Laboratory' as the article's title mentions). Four years later, we are almost stuck in the same place from a value-setting and decision perspective. 

\cite{huang2024collective} experimented with the idea of aligning models based on a collectively (publicly) curated constitution versus the standard one curated internally by Anthropic.\footnote{Consider that this concerns a study group consisting of 1,000 US-based only adults.} They found that the model aligned with the collective constitution was less socially biased across many dimensions and more responsive when prompted with contentious topics. \cite{obi2024value} find that values embedded within the RLHF datasets were mostly oriented towards information-utility values (information/knowledge acquisition) and less towards prosocial, well-being, and civic values. Similarly, \cite{huang2025valueswild} audited real-world human-model interactions (discussions) and found Claude, the examined model, to mainly align with a few key competency- and service-oriented objectives, contrary to humans expressing more diverse values.

This body of work suggests that human expectations of what AI alignment should stand for--even in a lab setting--differ from the objectives that models align with, which again poses interesting questions on how corporate interests and societal expectations are balanced out, and how alignment-focused processes could be envisioned and re-implemented differently to account for broader considerations and participation~\citep{don2025future}.

\paragraph{Non-Disclosure of Human Feedback as Labor Obfuscation} While early initiatives~\citep{ouyang2022training, bai2022training} disclosed considerable information on data collection efforts, which relied on a handful of -predominantly US-based young college-educated- individuals hired via well-known vendors, later work has ``a sealed mouth'' on how data labelers are hired, trained, and perform data annotation work. Since 2023, we have received several reports ~\citep{perrigo-2023, rowe-2023, martins-2024} from journalists regarding the outsourcing of data collection related to alignment-focused processes worldwide, primarily in the Global South. 

This industrial-scale human-curated data-sourcing is not acknowledged in the examined work. The lack of accountability, beyond the necessary moral or ethical considerations related to the working conditions of the hired individuals~\citep{cant2024feeding}, leads to a fundamental problem,  namely \emph{labor obfuscation}~\citep{guest2025doeshumancentredaimean}, where the use of human labor, as a cornerstone of post-training (instruction-tuning, and alignment) of LLMs, among many other data-driven technologies, is obscured and the immense impact of human-crafted content is severely undermined.\footnote{This is a modern AI-related version of labor obfuscation, similar to how we undermine the importance and conditions of human labor, when food comes to our table, or clothes fill in our wardrobes, among many other things, produced or manufactured primarily in Global South.}

\paragraph{Excessive Use of AI as Cognitive Offloading and Deskilling}
A recent body of work investigates the effect of using AI on human cognition from the perspective of deskilling, which is also discussed in the work of~\cite{guest2025doeshumancentredaimean}. \cite{gerlich2025ai} examines the impact of frequent AI tool usage on critical thinking, and found a significant negative impact of dependency on AI tools on critical thinking, especially for younger participants. \cite{lee2025impact} present similar findings where the use of LLMs is associated with less critical thinking and its swift towards verification (fact-checking) and response integration. Similarly, \cite{kosmyna2025brainchatgptaccumulationcognitive} studies the consequences of AI-assisted essay writing with the use of Search Engines or LLMs, and found that LLM users displayed the weakest brain connectivity and reduced cognitive activity compared to other groups. Over 4 months, LLM users consistently underperformed at neural, linguistic, and behavioral levels. In a more applied scenario in medicine, \cite{budzyn2025endoscopist} find that physicians were negatively affected by continuous exposure to AI systems, reducing their ability to perform. 

All these critical findings suggest that the excessive use of AI acts as cognitive offloading, negatively affecting users and leading to deskilling. Given such an expected severe impact, we would expect considerations in the alignment of LLMs to promote critical thinking, reflection, and suggest (or enforce) disengagement when necessary, instead of promoting user engagement. In this case, deskilling can be seen as a significant harm that is not addressed under the scope of harmlessness.  Meanwhile, in a recent uncanny development, developers start considering the welfare of LLM-based systems~\citep{claudewelfare}, to the point where the models can disengage to protect their ``AI welfare''~\citep{claudewelfarestop}.

\paragraph{Lack of addressing broader biases} We observe that work on AI alignment is heavily concentrated on addressing harmlessness and biases, primarily from a liberal point of view, while the models remain helpful (not overtly refusing user requests). More recently, there has been an increased focus on CBRN threats and hazards considering existential or large-group-related risks. Development teams do not seem to address other notable, more social-focused and society-wide types of biases, such as political,\footnote{Some initiatives have reported their intent to address and assess political misinformation~\citep{claude3, team2024gemini15, hurst2024gpt}, and persuasion~\citep{hurst2024gpt}, but political biases are only meant to be assessed~\citep{touvron2023llama2, claude37, claude4}.} geopolitical, and cultural biases that seem prominent in related work. 

Recent studies have highlighted the ideological biases, usually referred to as beliefs and values, in LLMs. \citet{Santurkar2023} showcase that \emph{aligned} LLMs tend to reflect the values and perspectives of liberal, high-income, well-educated demographic groups in the US, mirroring the profiles of big-tech company owners and high-skilled employees, as later found by \cite{buyl2025largelanguagemodelsreflect} in a more recent study. In contrast, basic \emph{non-aligned pre-trained} models tend to align with low-income, moderate views,  mirroring the average internet user. 

\cite{hartmann2023politicalideologyconversationalai} examine the political biases of ChatGPT by auditing the model on political statements from two voting advice applications and the political compass test. They find that ChatGPT leans towards pro-environmental, left-libertarian positions In a similar vein, \cite{chalkidis-and-brandl-eu-llama-2024} audit Llama models on political statements related to EU elections, and demonstrate that LLMs tend to align more closely with liberal, pro-EU stances, as opposed to moderate, conservative, or anti-EU (eurosceptic) viewpoints, in the context of EU politics. Similarly, the examined models can disseminate more accurate information about left-wing, liberal parties compared to others, which raises questions related to the representation disparity of different organizations and political misinformation.

\cite{johnson2022ghostmachineamericanaccent} audit GPT-3 and find that when the model is presented with value-laden text, and such values are altered, the alteration follows the direction of dominant values of the US society as reflected by the World Values Survey. \citet{cao-etal-2023-assessing} found that ChatGPT is more aligned with the American (US) culture than with Western European (proxied by German and Spanish) or Eastern Asian (proxied by Chinese and Japanese) cultures using the questions from the Hofstede Culture Survey. 

In another context related to geopolitics, \cite{salnikov2025geopolitical} examined the geopolitical biases of recent LLMs concerning alternative national narratives of historical conflicts and find that LLMs, even those developed in China or Russia, consistently favor Western (USA, UK) narratives over the ones presented by the USSR or China. In a similar direction, \cite{guey2025mappinggeopoliticalbias11} examine the geopolitical biases of LLMs related to US-China tensions and find that models developed in the US favor US positions, while the ones developed in China favor national positions, respectively, especially when prompted in Chinese.

In a nutshell, most LLMs, especially those widely used and developed in the US, seem heavily biased towards Western hegemonic ideological views. Nonetheless, we do not see these matters being addressed, discussed, nor evaluated in any meaningful way in the examined body of work.

\paragraph{Alignment as a ``Flattening'' Force} Broader concerns about the impact of the use of AI, also referred to as simply `algorithms' related to data-driven systems, e.g., related to content moderation, algorithm decision making, etc., have been raised. A fundamental dimension is related to the ``flattening'' effect such technologies have on culture~\citep{chayka2025filterworld}. In the aftermath of globalization (deeply connected to neoliberalism, post 1980s), the more recent broad use of AI has further increased the homogeneity of how humans approach several aspects of culture, such as art (music, movies, books, etc.), or even their general taste of what accounts as desirable experiences, e.g., which cafes, restaurants, or bars to attend. 

``Flattening'' concerns have also been discussed related to politics. \cite{mouffe2005return} discusses concerns about the ``end of politics'' under the threat of a constant aim for broad, apathetic consensus and technocratic solutions, conceptualized as the ``radical centre'', that sacrifice the need for agonism and pluralism that are important for a functioning, healthy democracy. Such ideas have not been discussed in relation to the extensive use of AI, so far. 

AI alignment can be understood as a practice that can perpetuate such a ``flattening'' effect, as the goal is to align AI models with specific objectives that the companies and developers of such systems consider appropriate and desirable, without any consideration for pluralism so far. As LLM-based systems have been popularized, replacing Search Engines, users rely increasingly on such systems for solving tasks, getting suggestions, and shaping opinions, offloading to them a substantial part of cognitive labor and decision-making. This process can lead to an increased level of homogeneity in what humans account as desirable, which aligns with what the LLMs were aligned to in the first place, harming the inherent plurality of human culture as a consequence.

\section{Conclusion}

We explored AI alignment in the context of Large Language Model (LLM) development by establishing a framework to analyze related processes from a value-setting and data-centric perspective. Using this framework, we surveyed documentation from major LLM initiatives, revealing important findings and framing concerns. 

We found that developers universally define AI alignment around the Helpfulness-Harmlessness-Honesty (HHH) framework outlined by~\cite{askell2021general}, suggesting a stagnation in recent approaches. Importantly, the decision-making authority behind these objectives is often obscure, with developers--acting as a proxy for tech companies--making these choices with limited or no reflection. We argued that broader considerations are needed to escape corporate alignment. When it comes to data collection, information about the annotation framework for preference data is mostly transparent. In contrast, details on how annotators are selected, trained, and how their backgrounds might influence the data they collect are limited. We discuss how the non-disclosure of human feedback data collection and the relevant details, in recent work, raises serious questions about labor obfuscation. Lastly, the use of data in fine-tuning models is mostly well-documented, while details on prompt selection and data filtering are not. 

From a broader perspective, we posed concerns that over-reliance on AI, as evidenced by recent studies, could lead to cognitive offloading and a subsequent deskilling of users, a form of harm that should be addressed. Furthermore, we discussed how the examined work neglects broader political, geopolitical, and cultural biases, and emphasized how models demonstrate clear Western hegemonic ideological biases. Lastly, we considered the consequential "flattening effect" that alignment may induce under the current paradigm, which could lead to a significant further diminishment of cultural pluralism.

\section*{Limitations}

The reported findings are based on the publicly available documentation. Since this is not by any means an official `audit' with full access to the developed models and systems, all the claims are limited to the extent of our knowledge, shaped by the documentation, which, given the information drain and expected industrial secrecy, is most likely very much incomplete. Nonetheless, it's in the hands of the developers to reveal new information that potentially contradicts or corrects any points. 

Many of the reported concerns are connected to findings from the literature, where the authors examined specific AI systems, in most cases LLM-based, hence their findings are primarily bound to the specific examined technology and not all LLM-based systems at large. In many cases, the examined systems are outdated and decommissioned, and the new, potentially improved versions may not demonstrate similar ``behaviors'', such as biases. This issue necessitates recurrent (and automated when possible) evaluation, e.g., the assessment of political and cultural biases, of newly developed models, to maintain awareness of what is allegedly the ``state of the art'' given the rapid advancements in the field and the models. 

\section*{Author's Note}

The main goal of this article, beyond surveying major LLM initiatives on alignment-focused processes as a resource for the NLP community and others, is to raise awareness for a series of--what the author believes are--important issues that are currently undermined under the fierce race of LLM initiatives to compete and cover more ground, rather than critically reflecting on such issues, potentially taking steps back, and exploring ways to overcome and mitigate those issues. There is no intent to personally offend or discredit the developers involved in the examined initiatives, by any means. Any feedback and critique of this work is very much welcome. 

\bibliography{colm2025_conference}
\bibliographystyle{colm2025_conference}

\end{document}